\documentclass{article}
\usepackage{spconf,graphicx}
\usepackage{amsmath,amssymb,amsfonts}
\usepackage{algorithm}
\usepackage[noend]{algpseudocode}
\usepackage{textcomp}
\usepackage{xcolor}
\usepackage{soul}
\usepackage[utf8]{inputenc}
\usepackage{katsuki}
\usepackage{subfig}
\usepackage{array}
\usepackage{bm}
\usepackage{booktabs}
\usepackage{url}

\title{Learning to Estimate Driver Drowsiness from Car Acceleration Sensors\\using Weakly Labeled Data}
\name{Takayuki Katsuki, Kun Zhao, Takayuki Yoshizumi}
\address{IBM Research - Tokyo, Tokyo, Japan\\
e-mail: \{kats, kunzhao, and yszm\}@jp.ibm.com
}

\begin{document}
%
\maketitle
\begin{abstract}
This paper addresses the learning task of estimating driver drowsiness from the signals of car acceleration sensors. Since even drivers themselves cannot perceive their own drowsiness in a timely manner unless they use burdensome invasive sensors, obtaining labeled training data for each timestamp is not a realistic goal. To deal with this difficulty, we formulate the task as a weakly supervised learning. We only need to add labels for each complete trip, not for every timestamp independently. By assuming that some aspects of driver drowsiness increase over time due to tiredness, we formulate an algorithm that can learn from such weakly labeled data. We derive a scalable stochastic optimization method as a way of implementing the algorithm. Numerical experiments on real driving datasets demonstrate the advantages of our algorithm against baseline methods.
\end{abstract}
\renewcommand{\thefootnote}{\fnsymbol{footnote}}
\footnote[0]{\copyright 2020~IEEE. Personal use of this material is permitted. Permission from IEEE must be obtained for all other uses, in any current or future media, including reprinting/republishing this material for advertising or promotional purposes, creating new collective works, for resale or redistribution to servers or lists, or reuse of any copyrighted component of this work in other works.~~DOI 10.1109/ICASSP40776.2020.9053100}
\renewcommand{\thefootnote}{\arabic{footnote}}
\section{Introduction}
\label{sec:intro}
\begin{figure}[tb]
    \centering
    \includegraphics[width=80mm]{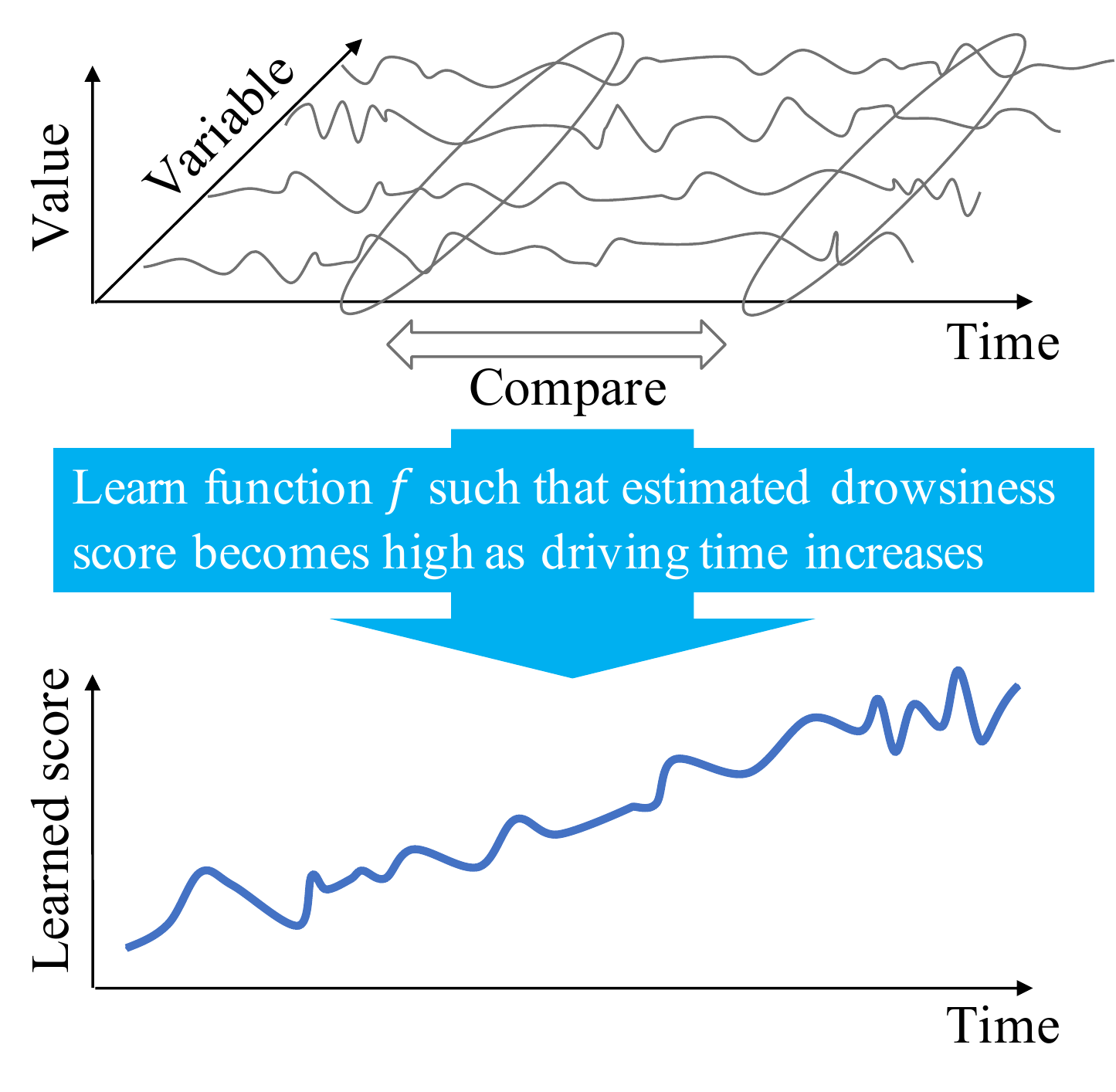}
    \caption{Learning to estimate drowsiness of drivers from car acceleration sensor data. The estimation function is learned such that the estimated drowsiness becomes large as the driving time increases.
    }
    \label{FigProblem}
\end{figure}
In this paper, we address the learning task of estimating driver {\it drowsiness}~\cite{friedrichs2010drowsiness,friedrichs2010camera}, a cause of numerous severe accidents. Such driver monitoring scenarios are typical embodiments of signal processing leveraging machine learning technologies~\cite{gideon2018multi,tour2018driver,regani2019car}. We take a light-weight and non-intrusive approach that does not require any image processing or invasive sensors and instead only requires acceleration sensors that capture anomalous acceleration, braking, and steering data reflecting the drowsiness of the driver. Such acceleration sensors can be found in the form of, e.g., smartphones and drive recorders. As our method does not use images, it entails less communication quantity and can be easily implemented as a cloud-based service provided over a cellular network~\cite{cvi2019}.

Regarding how the labeled training data are obtained, it is not easy for drivers themselves to add labels in a timely manner when they themselves are drowsy. Instead, equipment, such as electroencephalogram (EEG)~\cite{friedrichs2010drowsiness} could be used, but their invasiveness and intrusiveness would non-negligible costs imposed on the drivers. Such devices are likely to bother drivers and may even be causes of anomalous driving. The obtained labels and data may thus be unnatural and inhibit the generalization ability of the learned model.

We propose a more practical learning approach to address the difficulty of labeling drowsiness. The key idea is to use {\it weakly} labeled training data. More precisely, we do not add a label (drowsy or normal) at each point in time that the driver is at the wheel but rather assign a label to the whole trip. We assume that this sort of labeling can be done by drivers, because most drivers can more accurately answer whether they felt drowsy while they were driving instead of having to specify exactly when they feel or felt drowsy during the trip. However, despite that it is much easier than adding labels to every point in time independently, such weak labels are likely to make the subsequent learning task more difficult. To deal with this difficulty, we impose two reasonable assumptions, as follows.
\begin{description}
    \item[Drowsiness should be measured on an ordinal scale]~\\
    Drowsiness is measured on an ordinal scale~\cite{johns1991new}. Although we cannot define the absolute magnitude, ratio scale, or interval scale of drowsiness, we can consider which of two different data points is the drowsier.
    \item[Drowsiness increases as the trip gets longer]~\\
    Drowsiness is at least partly due to tiredness, which usually monotonically increases over time during a trip. It is thus logical to assume that drowsiness also increases over time during a trip labeled drowsy.
\end{description}

On the basis of the above assumptions, we learn a function that estimates a drowsiness score at each timestamp. From the first assumption, the function should be learned so as to preserve an order relation within its estimated drowsiness scores. In particular, we designed a learning algorithm based on a loss function comparing a pair of estimated scores of sampled data at once. Regarding the second assumption, the loss assumes that a driver is less drowsy earlier in a trip and more drowsy later in it, as shown in Fig.~\ref{FigProblem}.
If a trip is labeled normal, the proportion of time the driver felt drowsy is significantly lower than that of trip labeled drowsy. Thus, normal trips will be less informative for learning and we decided to learn the function using only drowsy trips for efficiency.
We derived a scalable stochastic optimization method for implementing the algorithm and empirically evaluated the effectiveness of the algorithm in numerical experiments using real driving datasets.

\section{Related Work}
There are two standard approaches to addressing the difficulty of labeling drowsiness: 1) using special equipment and 2) detecting anomalous driving without using labeled data.

Most studies taking the first approach used invasive and/or intrusive equipments to capture their data, which include physiological sensors, such as EEG and Electrooculogram (EOG)~\cite{friedrichs2010drowsiness,friedrichs2010camera,Lawoyin2014drowsiness}, image sensors~\cite{friedrichs2010camera,rathod2018camera}, and/or the managed simulated experiments~\cite{Lawoyin2014drowsiness}. By using labeled data gathered in these ways, we can learn a classifier for drowsy driving and use it for validation; however, the invasiveness and intrusiveness of these methods are problematic for the drivers.

The second approach is to formulate the task as a detection of anomalous data~\cite{ide2009proximity,egilmez2014spectral}, which requires no labeled training data; here, we might define the detected anomalous data as indicating drowsiness. However, in this case, many anomalies would be irrelevant to making a decision about drowsy driving, and such a method would detect {\it any kind of anomaly}~\cite{zheng2016unsupervised,chiou2016abnormal,nirmali2017vehicular}, indicating, e.g., some other form of dangerous driving, equipment failure, and changes in the environment (road surface and weather). It is non-trivial to extract only data indicating drowsy driving from all such anomalies.

Additionally, we can naively take a third approach that uses our weakly labeled training data, one in which we regard drowsy and normal trips as positive and negative samples respectively and apply a binary classification algorithm such as logistic regression.
In our experiment, therefore, we compared our proposed method with the second approach (anomaly detection) and the third approach (classification into drowsy and normal) that can be trained on our weakly supervised setting.

\section{Drowsiness Estimation via Learning from Weakly Labeled Data}
\label{sec:learning}
We formulate the estimation task for driver drowsiness as a machine learning problem. Our goal is to construct a model for estimating a drowsiness score, $y \in \mathbb{R}$, from $D$-dimensional features, $\bx \in \mathbb{R}^D (D\!\in\!\mathbb{N})$. $\bx$ are computed from observation signals of car acceleration sensors at each timestamp to ensure a fast response and efficient use of memory.

We learn an estimation function $f$ that takes an observed $\bx$ as input and computes an estimated score for $y$, $\hy$. The optimal estimation function $f^*$ is given by
\begin{align}
    \label{objective}
    f^* \equiv \argmin_{f} \mathcal{L}(f),
\end{align}
where $\mathcal{L}(f)$ is the expected loss when the estimation function $f$ is applied to samples $\bx$ distributed in accordance with an underlying probability distribution for a drowsy trip, $p(\bx)$. $\mathcal{L}(f)$ is defined using the expectation $E$ over $p(\bx)$ and the loss function $L(f)$:
\begin{align}
    \label{expectedLoss}
    \mathcal{L}(f) &\equiv E[L(f)].
\end{align}

From the assumptions stated in Introduction, we design the loss function $L(f)$ by using an ordinal scale that allows us to compare a pair of data. Our loss function takes a pair of i.i.d. samples from $p(\bx)$ at once as input and return a larger loss when the drowsiness score estimated for the sample with shorter driving time (older timestamp) is greater than that with the longer driving time (newer timestamp). Using samples $\bx_t$ and $\bx_u$, where the sub-indices $t$ and $u$ represent timestamps, the following loss function satisfies this condition:
\begin{align}
    \label{loss}
    L(f; \bx_t, \bx_u)\equiv \mathrm{max}\big(0,1-\mathrm{sgn}\big(t-u\big)\big(f(\bx_t) - f(\bx_u)\big)\big),
\end{align}
where $\mathrm{sgn}(\bullet)$ is a sign function.
Our loss function resembles ones for ranking or ordinal regression~\cite{cohen1998learning,burges2005learning,sculleylarge}. The key difference is to consider the driving time as the ground truth for the optimal order.
To learn $f$, we need to minimize the objective function (Eq.~\eqref{objective}) with the loss (Eq.~\eqref{loss}). In the following subsection, we derive a scalable implementation by using an empirical approximation and stochastic optimization.

\subsection{Stochastic Optimization for Scalable Learning}
\begin{algorithm}[t]
\caption{Stochastic optimization for learning to estimate driver drowsiness.}
\label{alg1}
\begin{algorithmic}[1]
    \Require Training data $\big\{\bX^{(\!i\!)} \big\}_{i=1}^{N}$ and
    hyperparameter $\lambda$
    \Ensure Model parameter $\btheta$ for $f$
    \State Let $\mathcal{A}$ be an external stochastic optimization method
    \While{No stopping criterion has been met}
    \State Randomly select $i$ from $1$ to $N$
    \State Randomly select $t$ and $u$ from $1$ to $T_i$ ($t \neq u$)
    \State $G \leftarrow \frac{\partial L\left(f; \bx_t^{(\!i\!)}, \bx_u^{(\!i\!)}\right)}{\partial \btheta} + \lambda \frac{\partial R(f)}{\partial \btheta}$
    \State Update $\btheta$ by $\mathcal{A}$ with the gradient $G$
    \EndWhile
\end{algorithmic}
\end{algorithm}
The expectation $E$ in Eq.~\eqref{expectedLoss} can be approximated using the sample averages. Let $\big\{\bX^{(\!i\!)} \big\}_{i=1}^{N}$ be $N$ trips labeled drowsy, where $\bX^{(\!i\!)}$ has $T_i$ samples as $\bX^{(\!i\!)} \equiv \big\{\bx_{\tau}^{(\!i\!)} \big\}_{\tau=1}^{T_i}$, and $\tau$ represents an index and also timestamp. Since comparing the drowsinesses for samples between different trips does not make sense, we calculate the loss with Eq.~\eqref{loss} from different samples in the same trip. Let $\bP^{(\!i\!)}$ be all candidate pairs of samples in $i$-th trip $\bX^{(\!i\!)}$; we empirically approximate the expectation in Eq.~\eqref{expectedLoss}, as
\begin{align}
    \label{empiricalLoss}
    &\mathcal{L}(f) \simeq \hat{\mathcal{L}}(f)\\\nonumber
    &~~~~~~~~\equiv \frac{1}{N}\sum_{i=1}^N \frac{1}{|\bP^{\{i\}}|}\sum_{\bx_t^{(\!i\!)},\bx_u^{(\!i\!)} \in \bP^{\{i\}}} L\Big(f; \bx_t^{(\!i\!)}, \bx_u^{(\!i\!)}\Big),\\
    &\mathrm{where}~~\bP^{(\!i\!)} \equiv \big\{\bx_t^{(\!i\!)}, \bx_u^{(\!i\!)}\big|\bx_t^{(\!i\!)} \in \bX^{(\!i\!)}, \bx_u^{(\!i\!)} \in \bX^{\!(i\!)} \big\}~~\mathrm{and}\nonumber
\end{align}
$|\bP^{\{i\}}|$ is the total number of candidate pairs in $\bP^{\{i\}}$.

For stable learning, we add a regularization term, $R(f)$, and then derive the gradient of Eq.~\eqref{empiricalLoss} as
\begin{align}
    \label{empiricalgradient}
    &\frac{\partial \hat{\mathcal{L}}(f)}{\partial \btheta} = \frac{1}{N}\sum_{i=1}^N \frac{1}{|\bP^{\{i\}}|}\sum_{\bx_t^{(\!i\!)},\bx_u^{(\!i\!)} \in \bP^{\{i\}}} \frac{\partial L\Big(f; \bx_t^{(\!i\!)}, \bx_u^{(\!i\!)}\Big)}{\partial \btheta}\nonumber\\
    &~~~~~~~~~~~~~~~~+ \lambda \frac{\partial R(f)}{\partial \btheta},
\end{align}
where $\btheta$ is the parameter vector of $f$ and $\lambda \geq 0$ is a regularization parameter that can be optimized on the basis of the mean error computed by cross-validation in training.

Algorithm~\ref{alg1} describes a stochastic optimization algorithm~\cite{sculleylarge} based on the gradient in Eq.~\eqref{empiricalgradient}, where we can use any stochastic gradient descent (SGD) method, such as Adam~\cite{kingma2014adam}, FOBOS~\cite{duchi2009efficient}, and Pegasos~\cite{shalev2011pegasos}, as the external stochastic optimization method $\mathcal{A}$.

By using the learned function $\hat{f}\equiv \argmin_{f} \hat{\mathcal{L}}(f)$ with Eq.~\eqref{empiricalgradient}, we can estimate $\hy$ for the new data as $\hy = \hat{f} (\bx)$.

For $f$ in the following experiments, we used a linear model, $\btheta^\top \bx$, where $\top$ denotes the transpose. For the feature vector $\bx$, we used the raw output of the sensors and additional features calculated from the raw output. The raw output consisted of three-axis accelerations (X, Y, and Z), the magnitude of the acceleration vector, speed, and direction at timestamps $t$ and $t\!-\!1$. The X, Y, and Z accelerations were aligned with the longitudinal axis of the vehicle (a positive value reflects acceleration, a negative value braking), the lateral axis (reflects turning), and vertical axis (reflects the road surface), respectively. The additional features consisted of the time derivatives of the raw output and anomaly scores of the features, where the anomaly scores were computed in the method reported in~\cite{ide2009proximity}. We used L$2$-regularization for the regularization term, $R(f)\equiv \|\btheta\|^2$.

\section{Experimental Results}
\label{sec:ex}
\begin{table}[t]
\caption{Comparison of proposed method and baseline methods in terms of AUC (larger is better) on real-world driving dataset. The best methods are in bold.
}
\label{ResultRD}
\centering
\begin{tabular}{cccc}
\toprule
\multicolumn{2}{c}{Method}&AUC$1$&AUC$2$\\
\midrule
Anomaly&Lasso&$0.41$&$0.51$\\
detection&Glasso&$0.36$&$0.44$\\
\midrule
&Logistic&$0.69$&$0.34$\\
Classification&SVM&$0.71$&$0.47$\\
&MLP&$0.79$&$0.59$\\
\midrule
\multicolumn{2}{c}{Proposed}&$\bm{0.82}$&$\bm{0.69}$\\
\bottomrule
\end{tabular}
\end{table}

We assessed the effectiveness of our approach in numerical experiments using real driving data collected at one sample per second ($1~\mathrm{Hz}$), a slow enough rate that data can be easily collected and handled by a cloud-based service provided over a cellular network~\cite{cvi2019} or by edge devices, such as smartphones and drive recorders. The driving data consisted of $11$ drowsy trips and $94$ normal trips, including highway and ordinary road trips. The average length of the trips was about $20$ minutes; the minimum was $86$ seconds, and the maximum was about $3$ hours. In total, there was about $40$ driving hours (over $100,000$ samples).

Since the score $\hy$ computed by the proposed method is represented on an ordinal scale of drowsiness, $\hy$ indicates how confidently drowsy driving is predicted. Thus, choosing a fixed decision threshold gives a classification rule for drowsy and normal data, and we can use it as a classifier. The experiment evaluated the classification performance of the proposed method on real driving data with regard to two types of the {\it area under the curve} (AUC): AUC$1$) classifying each trip as either a drowsy trip or normal trip based on the maximum score of $\hy$ during the trip and AUC$2$) classifying each sample with a drowsy timestamp or normal timestamp based on the raw score $\hy$. AUC can evaluate the goodness of the estimated score for classification with a continuously changing threshold. We used $11$-fold cross validation by using stratified sampling, where, for each fold of the validation, we sampled the same proportion, $\frac{1}{11}$, of trips from both drowsy and normal trips. In each of the validations with a total of $11$ drowsy trips, we chose $1$ drowsy trip and $\frac{1}{11}$ normal trips for testing and used the others for training.

\begin{figure}[t]
	\centering
	\includegraphics[width=75mm]{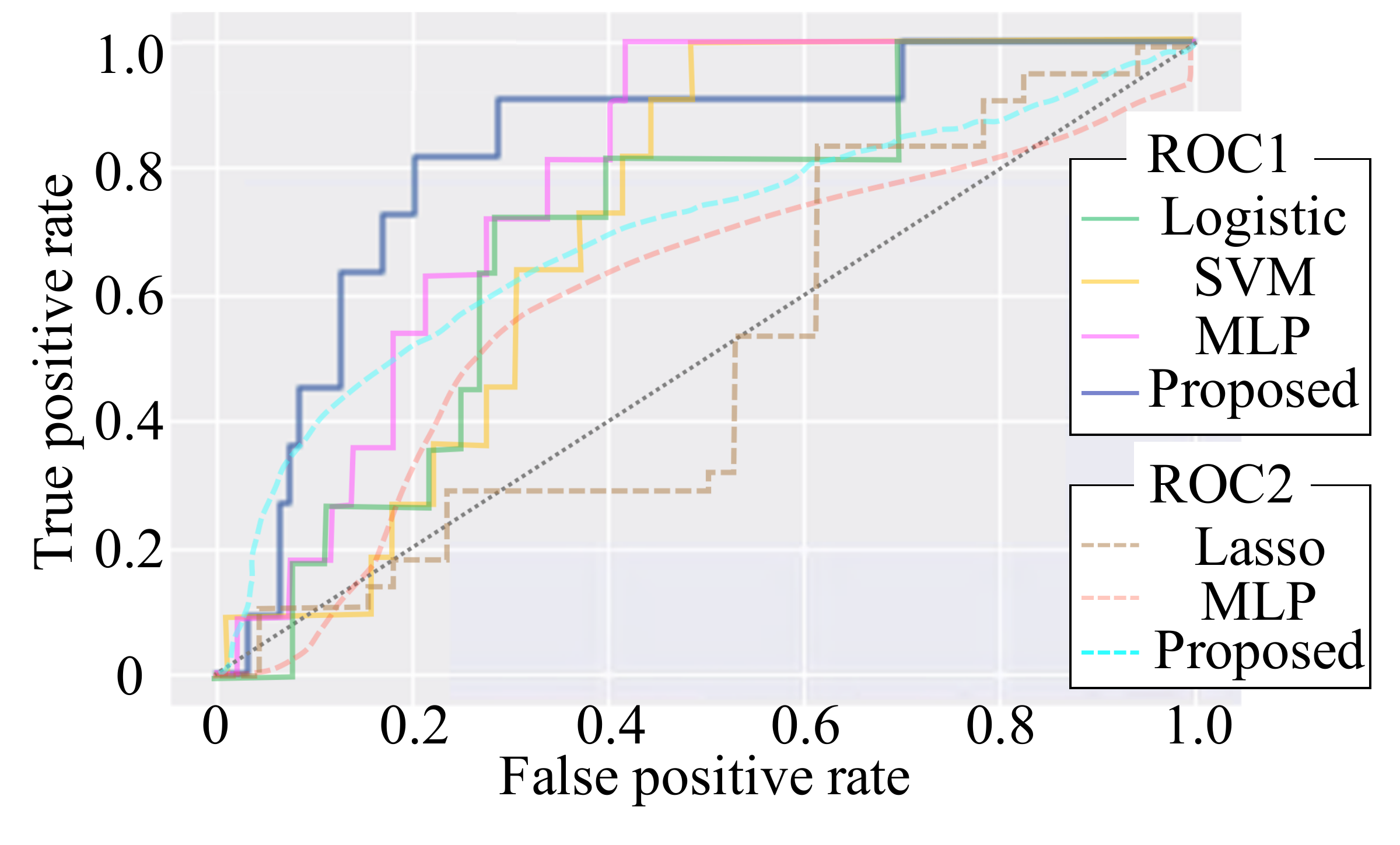}
	\caption{ROC curves comparing proposed method and baseline methods (depicted methods achieving AUC over $0.5$). Gray dotted line is plotted to represent random selection.}
	\label{FigRoc}
\end{figure}

Table~\ref{ResultRD} compares the results of the proposed method with two anomaly detection approaches and three classification approaches. As the anomaly detection approaches, we used Lasso~\cite{meinshausen2006high} and Graphical Lasso (Glasso)~\cite{ide2009proximity}, which are traditional but still state-of-the-art methods in anomaly detection for industrial sensors having linear correlations~\cite{hara2017consistent}. As the classification approaches, we used logistic regression with L$1$-regularization (Logistic), a support vector machine (SVM) with a radial basis function kernel, which was used in~\cite{Lawoyin2014drowsiness}, and a deep neural network (MLP: a $6$-layer multilayer perceptron with ReLU~\cite{nair2010rectified}, more specifically $D$-$100$-$100$-$100$-$100$-$1$), which was used in~\cite{friedrichs2010drowsiness,friedrichs2010camera}. The anomaly detection methods were learned using only normal trips; we used their anomaly score as the drowsiness score. The supervised classification methods were learned such that they would separate the samples from the drowsy trips and the normal trips, where we supposed that all of the samples in the drowsy trips belong to drowsy timestamps and those in the normal trips belong to normal timestamps. The proposed method was learned by using only drowsy trips. We used Glasso as the feature extraction (anomaly detection) model for the supervised methods and the proposed method.

From Table~\ref{ResultRD}, we can see that the performance of our method was better overall than those of the baselines. The AUCs for Lasso and Glasso were smaller than $0.5$, which was an unreasonable result. In general, AUC is expected to be larger than $0.5$ (random selection), but such results may occur with unsupervised anomaly detection methods because they detect all anomalies and those indicating drowsy driving are just some of them. AUC$2$ for Logistic and SVM was also smaller than $0.5$ because of severe underfitting in our weakly supervised setting, where samples were not correctly labeled drowsy or normal timestamp. MLP achieved good performance, close to that of the proposed method especially in AUC$1$ at the cost of an expensive computation for not only training but also prediction with a neural network. The computational complexities of the proposed method and MLP were both $\mathcal{O}(n)$, where $n$ is the number of parameters. Since the number of parameters for MLP was $100$ times larger than that for the proposed method, the computational complexity of MLP was $100$ times larger than ours. The actual computational time of MLP was about $30$ times longer than ours written in python, and it did not meet our light-weight requirement.
The receiver operating characteristic (ROC) curves used for computing AUC$1$ (ROC$1$) and AUC$2$ (ROC$2$) are depicted in Fig.~\ref{FigRoc}. The ROC curve represents the average relationship between the true positive rate and false positive rate for each method at various threshold settings. The curves of the proposed method were closest to the upper left corners in both comparisons on ROC$1$ and ROC$2$, which indicates that our method outperformed the baselines.

Table~\ref{ResultIF} lists the features that the proposed method calculated to have a large absolute weight. The X-jerk feature had the largest weight, which implies unstable acceleration and braking. Such unstable operation is likely to happen when a driver is drowsy. We also confirmed that the drivers actually felt tired or drowsy at the timestamps classified as drowsy driving by the proposed method.

\begin{table}[t]
\caption{Important features for estimating drowsiness.}
\label{ResultIF}
\centering
\begin{tabular}{cc}
\toprule
Important feature&Weight\\
\midrule
X-jerk&$0.073$\\
X-acceleration&$0.065$\\
Z-jerk&$0.048$\\
Direction-acceleration&$-0.038$\\
Magnitude of acceleration vector&$-0.044$\\
Y-acceleration&$-0.047$\\
\bottomrule
\end{tabular}
\end{table}

\section{Conclusion}
We formulated a learning problem for estimating driver drowsiness from car-acceleration sensor data using a weakly labeled data. The data for the whole trip rather than at each timestamp were labeled as drowsy or normal. We proposed a learning algorithm based on the assumption that some aspects of driver drowsiness increase over time due to tiredness. We developed a scalable stochastic optimization method for implementing the algorithm. An experimental evaluation on real driving datasets demonstrated that the proposed method outperformed the baseline methods.
Our algorithm requires only three-axis accelerations, speed, and direction data. It is thus a light-weight non-intrusive approach that can be installed in standard smartphones or drive recorders and does not place a burden on drivers.
The next step of our work is to complete a more exhaustive experiment using more real-world driving data. Exploring other practical loss functions with weakly labeled training data would be another interesting direction.

\fontsize{10.0pt}{10.0pt} \selectfont
\bibliographystyle{IEEEbib}
\bibliography{ref}

\end{document}